# TRANSLATION OF TELUGU-MARATHI AND VICE-VERSA USING RULE BASED MACHINE TRANSLATION

Dr. Siddhartha Ghosh[1], Sujata Thamke[2] and Kalyani U.R.S[3]

[1]Head of the Department of Computer Science & Engineering, KMIT, Narayanaguda, Hyderabad
siddhartha@kmit.in
[2]R&D Staff of Computer Science & Engineering, KMIT, Narayanaguda, Hyderabad
sujata.thamke@gmail.com
[3]R&D Staff of Information Technology, KMIT, Narayanaguda, Hyderabad
upadhyayula.kalyani@gmail.com

## ABSTRACT

*In today's digital world automated Machine Translation of one language to another has covered a long way to achieve different kinds of success stories. Whereas Babel Fish supports a good number of foreign languages and only Hindi from Indian languages, the Google Translator takes care of about 10 Indian languages. Though most of the Automated Machine Translation Systems are doing well but handling Indian languages needs a major care while handling the local proverbs/ idioms. Most of the Machine Translation system follows the direct translation approach while translating one Indian language to other. Our research at KMIT R&D Lab found that handling the local proverbs/idioms is not given enough attention by the earlier research work. This paper focuses on two of the majorly spoken Indian languages Marathi and Telugu, and translation between them. Handling proverbs and idioms of both the languages have been given a special care, and the research outcome shows a significant achievement in this direction.*

## 1. INTRODUCTION

Machine Translation(MT) is a sub-field of computational linguistics that investigates the use of software to translate text or speech from one natural language to another natural language. Machine Translation performs a simple translation of words from one natural language to another language, but that cannot produce a good translation of text i.e. recognition of whole phrases and their equivalent meaning should be present in the target language.

Machine Translation mentions the use of computers to convert some or the entire task of translation between human languages. Development of bilingual Machine Translation system for any two natural languages with electronic resources and tools is a challenging task. Many practices are being done to develop MT systems for different languages using rule-based and statistical-based approaches. Machine Translation systems are specially designed for two particular languages, called a bilingual system, and for more than a single pair of languages, known as multilingual system. A bilingual system may be either unidirectional, from one Source Language (SL) to Target Language (TL), or may be bidirectional. Multilingual systems are bidirectional, but most bilingual systems are unidirectional. Machine Translation methodologies are commonly categorized as direct, transfer, and Interlingua. The methodologies differ in the analysis of the SL and extent to reach a language independent representation of meaning between the source and target languages. Barriers in good quality Machine Translation output can be attributed to ambiguity in natural languages. Ambiguities are classified into two types: **structural ambiguity** and **lexical ambiguity.**

India is a linguistically rich area. It has 22 constitutional languages, which are written in 10 different scripts. Hindi is the official language of the Union. Many of the states have their own regional language, which is either Hindi or one of the other constitutional languages. In addition, English is very widely used for media, commerce, science and technology, and education only about 5% of the world's population speaks English as a first language. In such a situation, there is a large market for translation between English and the various Indian languages.

### 1.1 Telugu Language

Telugu is one of the major languages of India. It is a Dravidian language frequently spoken in the Indian state of Andhra Pradesh. There were 79 million speakers in 2013. In India Telugu language occupies third position which is been spoken by large number of native speakers.

### 1.2 Marathi Language

Marathi is an Indo-Aryan language .It is mainly spoken in Maharashtra. Marathi is one of the 23 official languages of India. There were 74.8 million speakers in 2013. In India Marathi language occupies fourth position which is been spoken by large number of native speakers.

## 2. RELATED WORKS

There has been a growing interest in Machine Translation. Machine Translation has been brought a great change in making the Indian language more flexible to learn and understand which has been brought into consideration by various translation techniques addressed for decades in the form of Language Translator. The well-known Parts Of Speech (POS) Tagging has been used to the two Indian Languages i.e. Telugu and Marathi where the dictionary has been created which consist of the text meaning as well as its POS, also proverbs/idioms which are difficult to retrieve the exact meaning in different Indian languages are been represented in a database which consist of the meaning of the proverbs/idioms. Retrieving the exact translated sentence is difficult but these are a small practice using direct translation.

## 3. PROBLEM DEFINITION

Indian Language Translation is one of the serious problems faced by Natural Language Processing where the proverbs/idioms is also one within them. To get the exact meaning of the word in different languages we should also know the grammatical arrangement of the sentences in every language. So we are working with Parts of Speech tagging for each and every word as we are working for Marathi to Telugu Translation and vice-versa. Getting the exact meaning of the proverbs/ idioms is difficult in Indian Languages because the meaning of the proverbs/idioms changes when the word to word translation takes place. The problem here is to resolve how to convert the source text to the target text without changing its meaning. So we have focused on two Indian Languages which have same grammatical arrangement of sentence.

## 4. MACHINE TRANSLATION APPROCAH

Generally, Machine Translation is classified into seven categories i.e. Rule-based, Statistical-based, Hybrid-based, Example-based, Knowledge-based, Principle-based, and online interactive based methods. The first three Machine Translation approaches are widely used. Research shows that there are fruitful attempts using all these approaches for the development of English to Indian languages as well as Indian languages to Indian languages. Figure.1, shows the classification of MT in Natural language Processing (NLP).

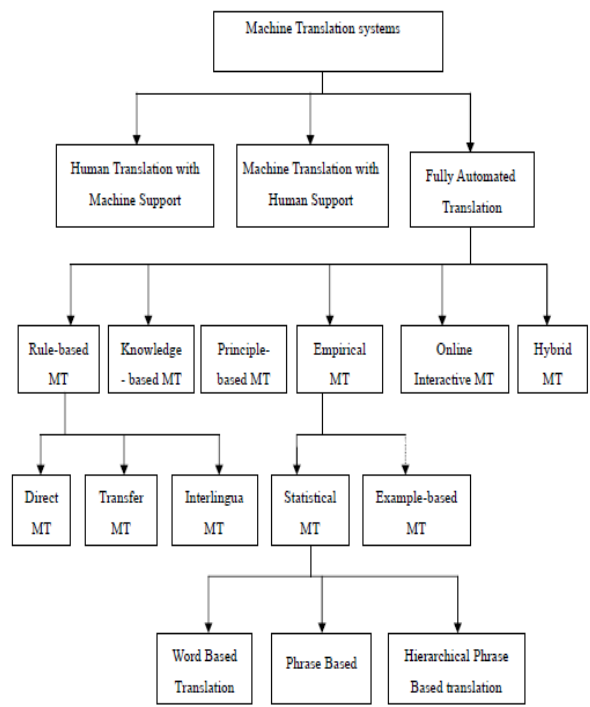

Figure. 1 Classification of Machine Translation

### 4.1 Rule-based Approach

The rule-based approach is the first strategy in Machine Translation that was developed. A Rule-Based Machine Translation (RBMT) system consists of collection of rules, called grammar rules, a bilingual or multilingual lexicon to process the rules. Rule Based Machine Translation approach requires a large human effort to code all of the linguistic resources, such as source side part-of-speech taggers and syntactic parsers, bilingual dictionaries. RBMT system is always extensible and maintainable. Rules play a major role in various stages of translation, such as syntactic processing, semantic interpretation, and contextual processing of language. Generally, rules are written with linguistic knowledge gathered from linguists. Transfer-based Machine Translation/Direct Translation, Interlingua Machine Translation, and dictionary-based Machine Translation are the three different approaches that come under the RBMT category. In the case of English to Indian languages and Indian language to Indian language MT systems, there have been fruitful attempts with all four approaches.

In this paper we have applied direct translation because both the Indian languages follow same sentence format i.e. SOV, so direct translation is applicable.

### 4.1.1 Direct Translation

In this method, the Source Language text is structurally analyzed up to the morphological level, and designed for a particular pair of source and target language. The performance of this system depends on the quality and quantity of the source-target language dictionaries, morphological analysis, text processing software, and word-by-word translation with minor grammatical adjustments on word order and morphology.

## 5. WORK CONTRIBUTED FOR TELUGU TO MARATHI TRANSLATION AND VICE-VERSA

To create a dictionary of Telugu and Marathi fonts in the database we have to go through the following process.

**1.** Download the appropriate Telugu http://telugu.changathi.com/Fonts.aspx) and Marathi (http://marathi.changathi.com/Fonts.aspx) fonts in your system then copy .ttf file and paste the file in the ***FONTS*** folder which is available in the ***Control Panel\Appearance and Personalization\Fonts.***

**2.** There are two ways to type the text in the Indian languages

**i)** This method is used by the people who know the keyboard formats of Telugu and Marathi fonts. After adding the fonts we need to go to Control Panel-→Clock, Language, and Region→region and languages→change keyboards or other input methods→general→installed services→add→Telugu (India) and Marathi (India)→ok.

**ii)** The following method is a simple method to type any Indian language.

To convert the words into appropriate Indian language we need to type the text in English which will directly convert the text into selected Indian language .To achieve this process we need to download ***Microsoft Indic Language Input Tool for Telugu*** (http://www.bhashaindia.com/ilit/Telugu.aspx install desktop version) and ***Microsoft Indic Language Input Tool for Marathi***(http://www.bhashaindia.com/ilit/Marathi.aspx install desktop version).

**3.** Here we have developed software for conversion by using .NET as front end and SQL as back end. The databases which are created are the collection of Telugu and Marathi words with its parts of speech. The following are the databases which are created:-

a) Telugu words and its Parts of Speech (POS).

b) Marathi words and its Parts of Speech (POS).

c) Telugu and its equivalent Marathi words.

d) Marathi and its equivalent Telugu words.

e) Proverbs/ Idioms in Telugu and Marathi.

f) Proverbs/ Idioms in Marathi and Telugu.

This paper deals with two Indian languages (Telugu and Marathi) which follows the same grammar rule i.e. SOV (Subject Object Verb) for both the languages so direct word to word translation can be possible only by providing source language and target language dictionaries.

Here we have concentrated on the POS Tagging. Basically rule formations are mainly depending upon the 'Morpho-syntactic' information's. With the use of these rules the Parts Of Speech Tagger helps us to add the appropriate POS Tags to each and every word. Our main goal is to translate the Marathi to Telugu and vice versa. The contribution of our work defines as follows.

## 6. PARTS OF SPEECH

Tagging means labelling. Parts Of Speech Tagging is the one where we add the Parts of Speech category to the word depending upon the context in a sentence. It is also known as Morpho-syntactic Tagging. Tagging is essential in Machine Translation to understand the Target Language. In NLP, POS Tagging is the major task. When the machine understands the TEXT then it is ready to do any NLP applications. For that the machine should understand each and every word with its meaning and POS. This is the main aim of our research. Particularly in MT when the system understand the POS of source language Text then only it will translate into target language without any errors. So POS Tagging plays an important role in NLP.

## 7. MACHINE TRANSLATION IN INDIAN LANGUAGES (TELUGU TO MARATHI-MARATHI TO TELUGU)

We have considered the two neighbouring states which are familiar with each other and the grammatical rule of both the languages are same by which the directly translation of the language takes place i.e. Maharashtra using Marathi as its mother tongue and Andhra Pradesh using Telugu as its mother tongue, both having same arrangement of words in the sentences i.e. SOV where the rearrangement of words doesn't takes place so direct translation is been considered. Let us take an example.

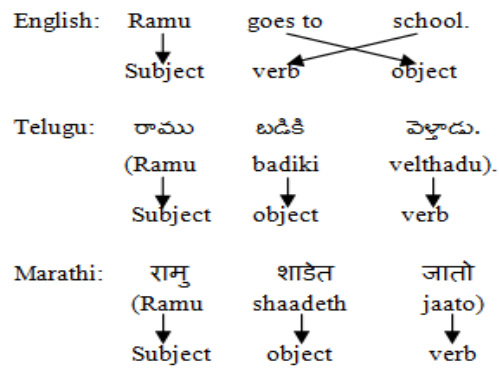

In English language the sentence formation follows Subject Verb Object where are in Telugu and Marathi the sentence formation is in the form of Subject Object Verb. As we are translating.

Telugu to Marathi both follows same format so direct translation has been done. Considering the above example where Ramu is the subject goes to is the verb and school is the object where as in Telugu and Marathi language Ramu/రాము/ रामु is the subject Badiki/బడికి/शाळेत is the object Velthadu/వెళ్తాడు/Jato जातो is the verb.

## 8. PROCEDURE FOR TRANSLATION

i) Select the language for translation then we need to enter the text in Telugu format with the help of Microsoft Indic Language Input Tool (Ref Figure.1).

ii) After entering the text, the sentence has been separated into words based on the delimiters (space, commas etc).The separated words are stored in an Array List.

iii) From the Array List each and every word will check with the database and gets its parts of speech for every word (Ref Figure. 2 and 3).

Consider the following example in this we have taken Telugu text which we want to convert it into Marathi. Let us consider an example; it shows how the translation has been done.

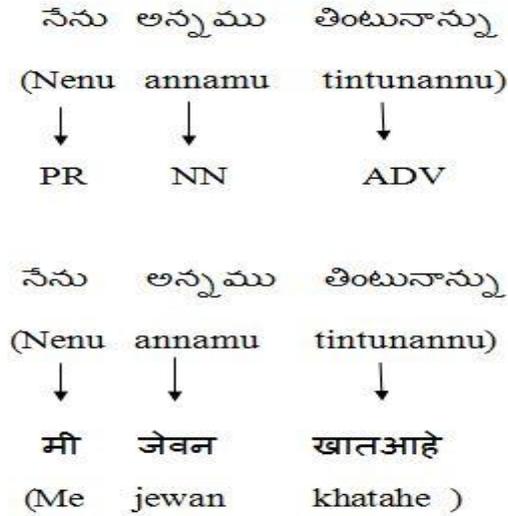

Here is the screenshots of the software which has been developed using Microsoft VisualStudio2010.To insert the Telugu and Marathi words in the database we need to keep the data type as **nvarchar** where n is used to support Multilanguage's i.e. Telugu and Marathi text in the database. Figure.2 shows the Home page of the developed software.

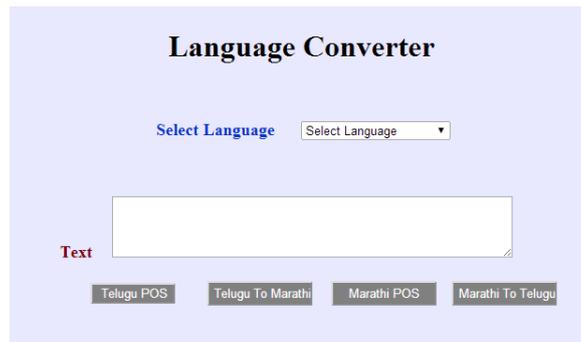

Figure. 2 Home Page

Figure. 3 shows the Parts of Speech for Telugu Text

Figure. 4 Shows Parts of Speech for given Marathi text

Converting Telugu Language into Marathi Language and Vice-Versa refer to Figure. 5 & 6 which shows how translation takes place with the help of database in the software which has been developed at our lab.

Figure. 5 shows the translation of given Telugu to Marathi Text

Figure. 6 shows the translation of given Marathi to Telugu Text

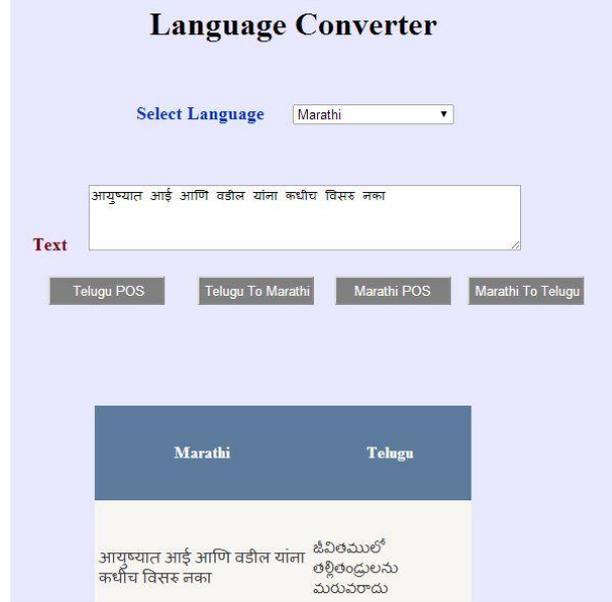

Figure. 7 shows the translation of Marathi proverb to Telugu proverb

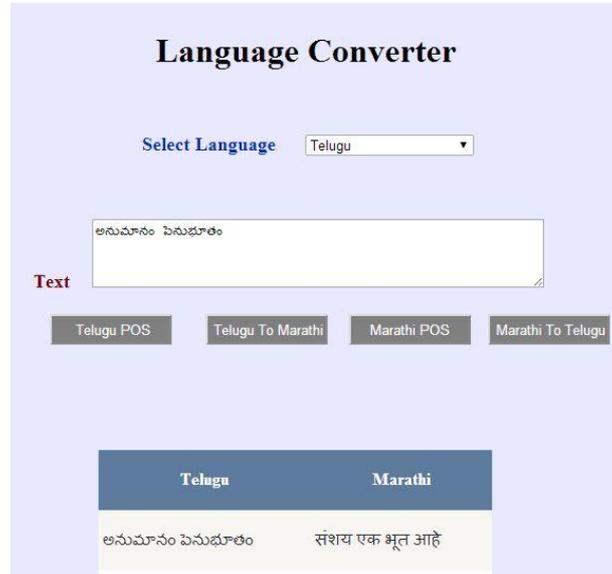

Figure. 8 shows the translation of Marathi proverb to Telugu proverb

## 9. PARTIAL PROVERBS LIST OF TELUGU AND MARATHI

అద్దం అబద్ధం చెప్పాదు
Addham abaddham cheppadhu.

అడిగే వాడికి చెప్పేవాడు లోకువ
Adigae vaadiki cheppaevaadu lokuva.

అదిగో పులి అంటే ఇదిగో తోక అంటారు
Adigo puli ante idigo thoka antaaru.

అందం అన్నం పెట్టాదు
Andham annam pettadu.

అత్త సొమ్ము అల్లుడు దానం
Attha sommu alludu daanam.

Figure. 9 List of Telugu proverbs

आधी विचार करा, मग कृती करा.
Adhi vichara kara, magh kruti kara.

आयुष्यात आई आणि वडील यांना कधीच विसरु नका.
Ayusyata aai ani vadila yanna kadhica visaru naka.

ज्याने स्वतःचं मन जिंकलं त्याने जग जिंकलं.
Jyane svatache mann jinkala tyane jaga jinkala.

जग प्रेमाने जिंकता येतं, शत्रुत्वाने नाही.
Jaga premane jinkata yete, satrutvane nahi.

राहायला नाही घर म्हणे लग्न कर.
Rahayala nahi ghara mhane lagna kara.

Figure. 10 List of Marathi proverbs

## 10. PROVERBS HANDLING ENGINE OF MACHINE TRANSLATION SYSTEM

In the case of proverbs, as the languages have its own proverbs which will not match with another language proverb so we have inserted the proverbs meaning in Telugu/Marathi proverb databases.

Figure. 11 Telugu to Marathi Proverbs Translation

Figure. 12 Marathi to Telugu Proverbs Translation

## 11. RESULT

We have observed that in Google translator if the word is not present then it is displaying the same word which is written in Telugu and also in the case of proverbs it is not converting the sentence in meaningful way and the conversion takes direct word to word translation. For example consider the below where we have given Telugu proverb as

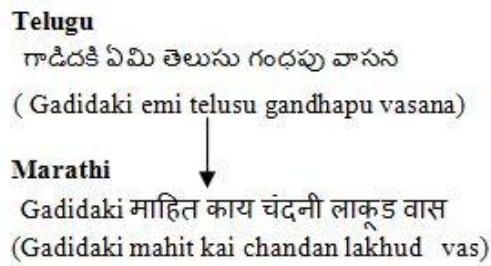

In the above example it is not changing "Gadidaki" (means donkey) into Marathi and the meaning of the proverb is also changing when the translation has been done.

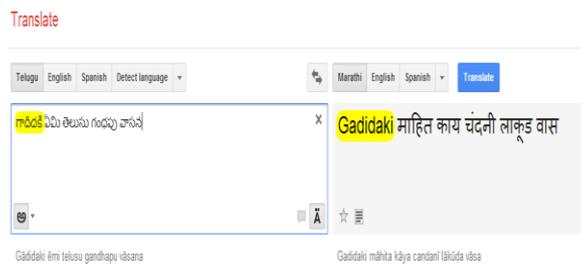

In case of BabelFish there are 14 languages present for translation out of that one Indian Language is present i.e. Hindi.

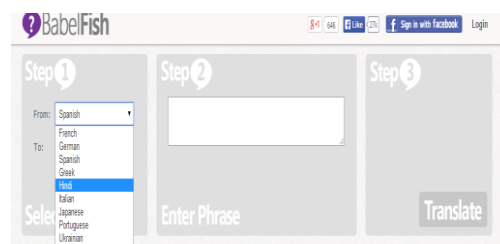

To overcome this problem we have added a separate dictionary for proverbs in which direct meaning of the proverbs has been inserted with its equivalent Telugu/Marathi text.

## 12. CONCLUSION

In this paper the translation of Indian Languages introduced and shown the parts of speech as well as the different Machine Translation approaches. Direct Translation was the technique to translate the Telugu and Marathi Language which have the same grammatical arrangement of sentence i.e. SOV. The Direct Approach can be possible only in some of the translation, but for more complex sentence the words are interchanged to get its proper meaning in the target language. This is especially true for real world problems where translation requires much complex algorithm to solve this problem when considered to the Indian Languages. Despite these already promising results, translation takes place using direct translation as well as and giving POS for each and every word, and proverbs/idioms are been solved by using dictionary which consist of meaning of the proverbs/idioms instead of word to word translation where there is no chance of wrong information. Also proverbs/idioms translation can most probably be improved. Further research might include a rule based approach, statistical approach for better translation. It is of great interest where Google translator as well as Babel Fish also failed to translate the exact meaning.

## ACKNOWLEDGEMENT

This work is supported by the Research & Development Department of under Computer Science Department of Keshav Memorial Institute of Technology, Hyderabad, India.

The author thanks to the director of institute to believe us and encourage us for the work. And also the reference papers which helped us a lot for our research.

**Dr. Siddhartha Ghosh** is the Head of the Department of Computer science & Engineering Branch in Keshav memorial Institute Of Technology, Hyderabad. He received his Ph.D and from the Osmania University, Hyderabad in Computer Science & Engineering. He is the best innovative faculty "smart city contest" winner of IBM in 2010. He is interested in the research areas of NLP, AI, Data Mining, Machine Learning, and Computing in Indian languages. He has about 24 national and international research publications. He has combindly authored two books published from USA.

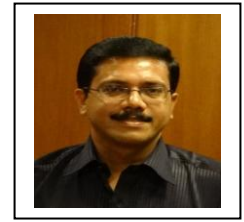

**Sujata M. Thamke is** working as a Research Assistant in Keshav memorial Institute Of Technology, Hyderabad. She received her Polytechnic from MSBTE and B.E from Amravati University in Computer Science & Engineering and Pursuing M.Tech in Computer Science & Engineering from Jawaharlal Nehru Technological University

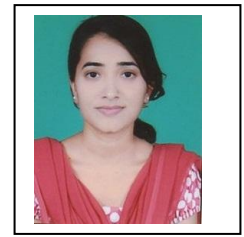

**Kalyani U.R.S** is a working as a Research Assistant in Keshav memorial Institute Of Technology, Hyderabad. She received her B.Tech degree in Information Technology From St.Mary's College of Engineering and Technology affiliated to Jawaharlal Nehru Technological University.

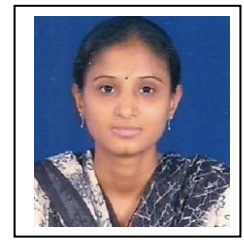